# Convolutional Neural Network Ensemble Learning for Hyperspectral Imaging-based Blackberry Fruit Ripeness Detection in Uncontrolled Farm Environment

Chollette C. Olisah[1]*, Ben Trewhella[2], Bo Li[3], Melvyn L. Smith[1], Benjamin Winstone[1], E. Charles Whitfield[4], Felicidad Fernández Fernández[4], Harriet Duncalfe[5],

[1]University of the West of England (UWE), Bristol, UK.
[2]Opposable Games Ltd, UK.
[3]Syngenta, Jealott's Hill Research Centre, UK.
[4]NIAB, East Malling, UK,
[5]Berry Gardens Growers, UK.

Chollette Olisah: *Correspondence to chollette.olisah@uwe.ac.uk, UWE, Coldharbour Ln, Stoke Gifford, Bristol BS16 1QY

**Abstract**

Fruit ripeness estimation models have for decades depended on spectral index features or colour-based features, such as mean, standard deviation, skewness, colour moments, and/or histograms for learning traits of fruit ripeness. Recently, few studies have explored the use of deep learning techniques to extract features from images of fruits with visible ripeness cues. However, the blackberry (Rubus fruticosus) fruit does not show obvious and reliable visible traits of ripeness when mature and therefore poses great difficulty to fruit pickers. The mature blackberry, to the human eye, is black before, during, and post-ripening. To address this engineering application challenge, this paper proposes a novel multi-input convolutional neural network (CNN) ensemble classifier for detecting subtle traits of ripeness in blackberry fruits. The multi-input CNN was created from a pre-trained visual geometry group 16-layer deep convolutional network (VGG16) model trained on the ImageNet dataset. The fully connected layers were optimized for learning traits of ripeness of mature blackberry fruits. The resulting model served as the base for building homogeneous ensemble learners that were ensemble using the stack generalization ensemble (SGE) framework. The input to the network is images acquired with a stereo sensor using visible and near-infrared (VIS-NIR) spectral filters at wavelengths of 700 nm and 770 nm. Through experiments, the proposed model achieved 95.1% accuracy on unseen sets and 90.2% accuracy with in-field conditions. Further experiments reveal that machine sensory is highly and positively correlated to human sensory over blackberry fruit skin texture.

# 1. Introduction

The blackberry fruit belongs to the class of soft fruits which is native to Europe, Asia, North, and South America[1]. The word 'soft' or 'hard' in relation to fruits implies the storage requirements and shelf life of the fruit. Soft[2] fruit examples, such as raspberry, blackberry, strawberry, and blueberry, usually require delicate handling and have a short shelf life compared to hard fruits[2] like banana, apple, mango, or dragon fruit. It is also worth noting that soft fruits are generally considered non-climacteric or ethylene insensitive, so it is not possible to pick them unripe and schedule their ripening after a period of storage like it is possible with hard fruits. In 2018, in the United Kingdom alone, the market value of soft fruits was estimated at £670 million[2], which indicates that there is an established and growing number of soft fruit consumers. Demand for soft fruit has risen, likely in part due to the promotion of eating healthy foods and the benefits of eating fruits rich in antioxidants (Zanini et al., 2015), and consumers expect excellent taste and quality. Typically, consumers associate the taste and quality of soft fruits with ripeness. Soft fruits, particularly blackberries, are best enjoyed when perfectly ripe. However, due to changes in growing conditions and seasons, and the ripeness traits hidden in a mature blackberry (Mikulic-Petkovsek et al., 2021), it is difficult to visually differentiate between a mature ripe berry from a mature unripe berry at the point of picking from the plant. During the picking process, pickers need to make the decision on the berry to pick within approximately three seconds by evaluating the colour appearance and the size of the drupelet. This impacts growers' ability to meet market needs and expectations. To measure the ripeness and overall quality of blackberries, many growers carry out sampling and targeted assessments on their harvested crop, testing samples for firmness, sweetness, and flavour (Cockerton et al., 2020), and/or a biochemical analysis to measure the contents of sugar and organic acids (Mikulic-Petkovsek et al., 2021). The former is based on subjective sensory evaluation, while the latter is based on objective laboratory assays, which are more accurate but quite labour-intensive. Both types of measures can only evaluate a small subset of the harvested crops and are destructive, thus adding to food waste.

Fruit ripeness estimation using non-destructive techniques is a well-researched area that has advanced from pure computational processes (Polder et al., 2002; Sinelli et al., 2008; Li et. al. 2018; Beghi et al., 2013; Shah et al., 2020) to processes enhanced by intelligence. For example, (El-Bendary et al., 2015; Hamza and Chtourou, 2018; Sabzi et al., 2019). Other examples with ready tools for in-field usage are (Cho et al. 2020; Gutierrez et al., 2019; Simonyan et al., 2014).

Computational processes of ripeness estimation include methods that investigate and analyse fruit ripeness using Vis-NIR imaging and have utilized spectral indices (features) from fruit images to estimate ripeness. This has been the line of study in fruit ripeness assessment for many years now. Polder et al., 2002 acquired images at 257 spectral bands which were averaged to obtain wavelengths of 396 nm and 736 nm corresponding to the ripeness stages of tomatoes. Using spectral features to train a linear discriminant analysis (LDA) model, they were able to show that spectral features reduced the classification error from 51% (with RGB features) to 19%. Sinelli et al. (2008) utilized near-infrared (NIR) spectroscopy to examine blueberry ripeness. Using partial least squares (PLS) regression algorithms, they showed that spectral features correlated with ripeness. Li et al. (2018) used two hyperspectral cameras covering the wavelength range between 600 and 1700 nm and applied multivariate statistical analysis to predict the fruit quality attributes including, firmness, colour, and soluble solid content which achieved good performance in fruit ripeness estimation. Beghi et al. (2013) acquired blueberry images in field settings using spectral bands from 445 nm to 970 nm. Using PCA, two spectral wavelengths, 680 nm, and 740 nm were identified to correlate with ripe and near ripeness, respectively. The spectral features were further used to create a blueberry ripeness index which was integrated into the design of a micro-controller-based handheld optical device for assessing ripeness level. Several other studies can be found in the review work by Shah et al. (2020).

---

[1] https://www.ecowatch.com/fruit-foragers-companion-2588139252.html
[2] https://the-secret-garden.net/soft-fruit-vs-hard-fruit-what-are-the-differeces-and-which-is-easier-to-grow.html

Ripeness estimation enhanced by artificial intelligence includes the fruit ripeness process researchers have approached by combining Vis-NIR or colour space conversion methods with machine learning. Here the discussion of the literature will progress from ripeness estimation of hard fruits to soft fruits. El-Bendary et al. (2015) converted RGB images of tomato fruit to the HSV colour space and extracted moment and histogram features which were further combined to form features, for discriminating between different levels of ripeness in tomato fruit. Using PCA, these features were downsampled and used as input to a support vector machine (SVM) classifier. The model achieved 84.30% and 90.80% on one-against-all ripeness levels and one-against-one ripeness levels, respectively. Hamza and Chtourou (2018) after applying a segmentation algorithm to separate apple fruit from the background image, extracted mean, variance, chromaticity, standard deviation, and skewness features which were trained using a multi-layered perceptron (MLP). Their method achieved a 98.33% accuracy on the test set. Sabzi et al. (2019) also applied a segmentation algorithm to separate foreground apples from background images but extracted colour-based features in hue saturation and luminance (HSL) channels which were fed as input to an MLP classifier that was optimized using genetic algorithms (GAs). The classifier was able to classify the apples into their respective classes: unripe, half-ripe, ripe, and overripe with an accuracy of 97.88%. A 93.02% classification accuracy was achieved (Castro et al., 2019) by combining colour space conversion and ML to cape gooseberry fruit classification. They extracted the mean values of RGB images as colour parameters and converted them to the L*a*b* colour space. Then, the colour features were combined with visual features for training on several ML algorithms, of which the SVM emerged as the best model. Steinbrener et al. (2019) captured images of different hard fruits using 16 spectral bands ranging from 470 to 630 nm, which were compressed into three-band images. These were further trained on the GoogleNet model that was pre-trained on RGB image data to form what they referred to as a kernel model. This model achieved accuracies of 96.55, 88.3%, 100%, 97.7%, and 100% for apples, avocados, bananas, mangoes, and tomatoes, respectively. Cho et al. (2020) argue that extracting the colour values in L*a*b* reveals firmness and translates to the ripeness of avocado fruit. The L*a*b* values were extracted from images of each avocado fruit collected with a smartphone camera and used as target values for their support vector regression (SVR) algorithm. After training the collected images, the SVR algorithm achieved a root mean squared error (RMSE) of 7.54. Worasawate et al. (2022) investigated the ability of ML algorithms to classify mangoes based on fruit appearance in terms of average colour and weight along with electrical data like capacitance. Of all the ML algorithms compared, the feed-forward artificial neural network (ANN) emerged as the best model, with an accuracy of 93.6% for a two-class (ripe, and unripe) classification experiment and 89.6% for a three-class (ripe, overripe, and unripe) classification experiment.

In (Worasawate et al., 2022) images of strawberries were acquired at 19 spectral bands from 405 nm to 970 nm and principal features were extracted using PCA. The resulting features were trained on several ML algorithms, and 100% accuracy was achieved with the SVM classification model. Kangune et al. (2019) trained a CNN classification algorithm on extracted contour and histogram features obtained from RGB, and HSV grapefruit images acquired in the field to classify each fruit as either ripe or unripe. Steinbrener et al., 2019 extended their study to include different soft fruits and achieved ripeness accuracies of 100% for both strawberries and grapes. Similarly, Gao et al. (2020) captured images of strawberry fruits using spectral bands ranging from 370 nm to 1015 nm. The spectral wavelengths of 530 nm and 528 were identified to clearly define bands of early ripe and ripe strawberries in the field and lab, respectively. Using PCA, they extracted three principal components and combined them as a three-channel image for detecting early ripe and ripe strawberries. These were further processed by training on an AlexNet model, pre-trained on the RGB ImageNet dataset, and achieved 98% and 100% for field and lab-collected images, respectively. In (Raj et al., 2022) they extracted spectral signatures related to strawberry fruit ripeness using a narrowband hyperspectral spectroradiometer. Using the spectral reflectance images as input to an SVM classifier, the proposed model was able to discriminate between ripe and unripe strawberries and achieved 98.2% accuracy.

Other research works such as (Saranya et al., 2021; Miragaia et al. 2021) applied DL algorithms directly to RGB images to estimate the ripeness levels of fruits. In (Saranya et al., 2021), a CNN architecture was used to classify bananas into ripe, partially ripe, ripen, and overripe classes and achieved a 96.14% accuracy on the test set. Miragaia et al. (2021) captured images of plum fruit in the field, cropped them to focus on the fruit of interest, and trained on the AlexNet model pre-trained on the ImageNet with modifications made to the dense layers. The proposed method in (Miragaia et al.

2021) was able to differentiate between three varieties of plums and classify them based on their levels of ripeness with a 92.83% accuracy.

Some studies have extracted spectral index features generated by averaging pixels in each segmented fruit region. Others extracted colour-based features, such as mean, standard deviation, and skewness values of intensity, colour moments, and/or histograms for training a classifier. The colour-based features are best established using laboratory-controlled images, given that stability is more likely to be achieved under such structured conditions rather than in field settings. The ripeness estimation models which used images captured in the visible range (Saranya et al., 2021; Miragaia et al. 2021; Raj et al., 2022) achieved good accuracies. Particularly, the high-performance accuracies of Vis-NIR images Gao et al. (2020), have shown they are better at overcoming the rigorous processes of extracting spectral features or visible colour-based features. Even though these lines of thought present a promising direction for fruit ripeness estimation, neither visible range captured images or colour-based features are suited to the challenges that mature blackberry fruit presents. The traits of ripeness in blackberry fruits at a mature stage are very subtle and cannot be discerned at visible range.

This paper proposes a non-destructive approach to in-field blackberry ripeness assessment by utilizing the capability of hyperspectral imaging and deep learning for detecting the traits of ripeness in a mature blackberry fruit under in-field conditions. The former involves the utilization of line scanning hyperspectral camera covering wavelengths ranging from 600 nm to 975 nm to experimentally determine the spectral wavelengths that best capture the reflectance characteristics of ripeness in a mature blackberry fruit. This is to be used for acquiring images at the determined wavelengths instead of exploring the spectral indices of ripeness. The latter comprises the design of a novel multi-input CNN ensemble classifier that is capable of learning the intrinsic ripeness characteristics of mature blackberry fruit images. The multi-input CNN ensemble model optimises the fully connected layers of the VGG16 architecture pre-trained on the ImageNet dataset and serves as a base for the ensemble learners. The homogeneous ensemble learning with stack generalization ensemble (SGE) framework is explored. To the best of our knowledge, this is the first study to investigate automated ripeness detection of blackberry fruits, and in a field environment. This paper's contributions are summarised as follows.

- Customize a stereo camera lens to filter light at wavelengths determined via analysis of acquired on-farm hyperspectral images.
- Design of a novel multi-input CNNvgg16 ensemble (MCE) architecture that adapts pre-trained weights to extract features that represent ripeness characteristics of mature blackberry fruits acquired at determined wavelengths. By optimising the fully connected layers of the network, their concatenated features are leveraged for building a homogenous ensemble learners' base model. Using a meta-learning algorithm within an SGE framework, the prediction results of each learner are leveraged for predicting the final class of a mature blackberry fruit, as ripe or unripe. The proposed model offers a new insight into fruit ripeness.

- Adopt a qualitative approach to statistically analyse the strength of correlation between the machine sensory and human sensory in fruit ripeness estimation using the Pearson correlation matrix.

The rest of this paper is organized as follows. The proposed methodology is presented in Section 2 and Section 3 is the experimental settings. Section 4 presents the experimental results and discussions while Section 6 concludes the paper.

## 2. Materials and Methods

### 2.1 Data collection

The experimental dataset consists of blackberry fruits acquired from two commercial blackberry farms located in the UK – one is the Windmill Hill Fruits Ltd in Herefordshire (52.0564° N, 2.7160° W), UK, while the other is the Clockhouse Farm in Kent (41.71°N 73.48°W), UK. In this paper, these farms are identified as Farm-A and Farm-B, respectively. To select the spectral wavelengths that correspond to ripeness levels of blackberry fruit, the categories of raw (0), unripe (1), near ripe (2), ripe (3), and overripe (4) blackberry fruits were sampled and imaged using the IMEC Ltd (Gent, Belgium) line scanning hyperspectral camera, which covered wavelengths from 600 nm to 975 nm, to image the sampled blackberry fruits at the spatial resolution of 2048xlength of scan and spectral resolution less than 10 nm. The camera was set up inside a stainless frame with six 12 Watts tungsten halogen lamps. Since this was a laboratory-driven experiment, the variability in spectral reflectance due to environmental factors (Hong, et al., 2018) remained unknown. It can be argued that if a fruit ripeness assessment model is capable of disentangling ripeness in an image of a blackberry fruit from constituent noise due to spectral variability, then the effect can be negligible. This is based on the fact that though the spectral variability introduces a decrease in reflectance spectra, the spectral wavelength remains the same as is seen in (Hong, et al., 2018). However, steps were taken to ensure only the blackberry fruit reflectance was considered. After imaging, the sampled blackberry fruits were segmented and averaged to generate the reflectance spectrum of blackberry ripeness. This spectrum averaging follows a process termed linear mixing. Figure 1 shows a significant difference in spectral reflectance in ripeness levels ranging from 600 nm to 850 nm with the most significant from 700 nm to 850 nm. The ripeness characteristics of mature blackberry fruits, near ripe (2) and ripe, show close reflectance characteristics over the spectral bands, even at near-infrared bands. However, a striking difference in the ripeness characteristics can be observed at a wavelength of 770 nm. Therefore, this paper explores the near-infrared spectral wavelength that best reflects ripeness characteristics and a visible spectral wavelength to maximize their respective cues for detecting ripeness in mature berries.

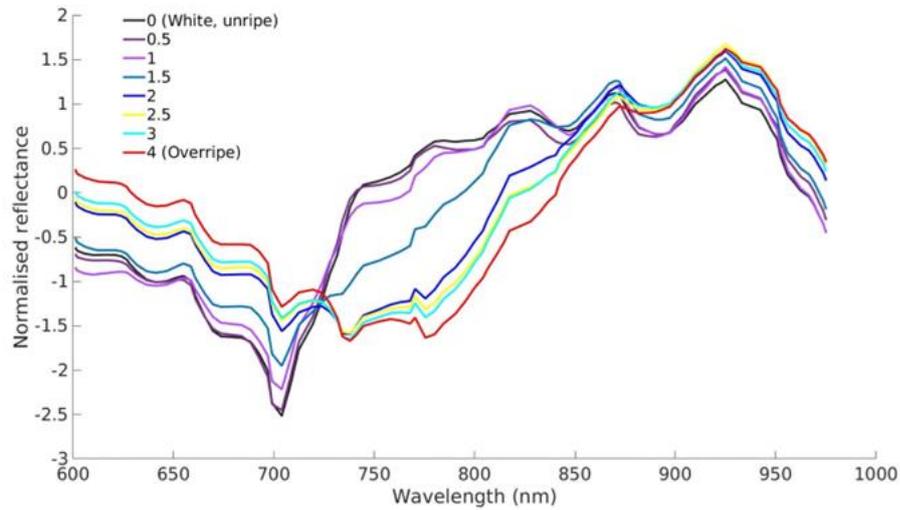

**Figure 1** Normalised reflectance spectra at different blackberry ripeness levels. This is best viewed in colour.

To acquire images in the Vis-NIR (bispectral images) wavelengths, an Arducam synchronized stereo 12MP camera was fitted with filters at spectral wavelengths of 700 nm and 770 nm, respectively. The two stereo cameras worked in synchrony, as can be seen in Figure 2. Figure 2 shows the use of the stereo camera on the farm, and its resulting image is a horizontally aligned image of the two wavelengths placed side-by-side to each other. Since the challenge for pickers is discerning between a mature ripe berry and a mature unripe berry or over-ripe berry, only the mature berries were captured. The mature ripe berry implies blackberry fruits that show the appearance and colour of a ripe berry and taste sweet, whereas the mature unripe berry implies blackberry fruits that show the same appearance and colour of a ripe berry but taste sour. In all, a total of 87 and 214 blackberry fruit samples were collected from Farm-A and Farm-B, respectively. Using the stereo camera and an RGB camera. It should be noted that the RGB camera is included in the collection of data for comparison purposes.

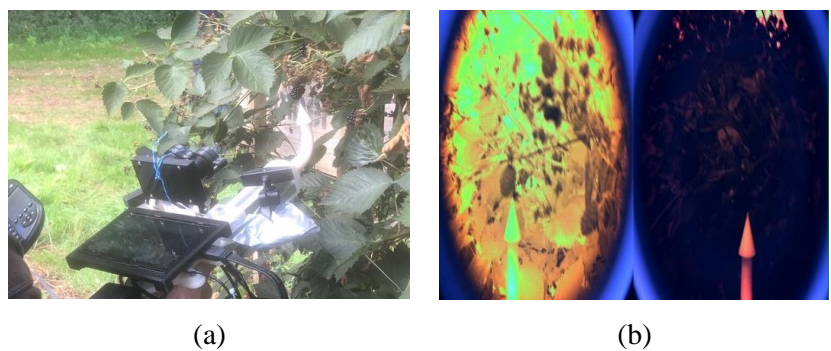

(a)          (b)

**Figure 2** On-farm blackberry imaging using multiple-camera mount with a pointer to aid fruit identification. a) stereo camera and RGB camera, b) a horizontally aligned image which is a sample of the stereo camera imaging.

## 2.2 Multi-Input CNN$_{vgg16}$ Ensemble Learner

A multi-input CNN$_{vgg16}$ ensemble learner (MCE) is proposed for classifying a blackberry fruit as ripe or unripe. This learner combines a multi-input CNN$_{vgg16}$ architecture and ensemble learning.

### 2.2.1 Multi-Input CNN$_{vgg16}$

The multi-input CNN$_{vgg16}$ architecture proposed in this work is structured on the strength of VGG16 (Simonyan and Zisserman, 2014) pre-trained on ImageNet. An illustration of the proposed multi-input CNN$_{vgg16}$ architecture deployed can be seen in Figure 3. The decision to use a pre-trained VGG16 model comes from its performance in comparison to other pre-trained models, such as DenseNet (Huang et al., 2017), Xception (Chollet, 2017), InceptionV3 (Szegedy et al., 2016), and MobileNet (Howard et al., 2017), which are usually chosen based on complexity and depth of the networks. As observed through the experimental results in Table 1 and Figure 5, the lesser the complexity and depth, the better the transferability from the base task to the target task. This is especially true for small sample datasets. A summary of modifications made to the networks is presented in Table 1.

As conventionally done in transfer learning, the learned weights generated at the convolutional layers from training VGG16 on the ImageNet dataset were frozen. Then through weight update at the fully connected (FC) layers, the network can learn the discriminative features of a mature ripe/unripe blackberry fruit. The VGG16 configuration (Simonyan and Zisserman, 2014) changes with 1) the type of input to the network – in this case, the input is a bispectral image acquired with camera filters set to spectral wavelengths of 700 nm and 770 nm. It has been shown that knowledge transfer from a domain of a different task and image format to another is feasible in transfer learning (Olisah and Smith 2019); 2) the size of the input – where each image is of size 32x32x3. The size of the input is critical to the network design because the bispectral images were generated using a modified stereo camera, where optical filters were added to each lens. The two lenses introduced a slight positional difference (known as the stereo disparity) between the images. The subsequent fuse of the two images in the spatial domain resulted in a distorted image. Also, conventional registration techniques were ineffective, which may also be due to the varying intensity of pixels between the images. Therefore, this problem was addressed by exploring the concatenation of features in the feature space with a reduced image size; 3) the number of neurons of the two FC layers – the number of neurons is dynamically determined through stratified k-fold cross-validation to determine the best FC layer that can learn the ripeness cues of the blackberry fruits at the two spectra, and 4) the best hyperparameters – where these are likewise determined through stratified k-fold cross-validation to result in the best configuration of the network. The rectified linear unit (ReLU) (Agarap, 2018) activation function is chosen for the FC layers because it is commonly used to overcome the vanishing gradient problem. Lastly, the classification layer is set to the sigmoid activation function to output values in the range [0,1] where values ≥ 0.5 indicate a mature unripe berry and values < 0.5 indicates a mature ripe berry. To avoid overfitting, during training of the multi-input $CNN_{vgg16}$, early stopping is combined with the model checkpoint. Overfitting results from the inability of the network to learn the relationship of within-class objects, but rather learns the noise in the data. Transfer learning can address the overfitting problem, but not completely because the transferability of features decreases with dissimilarity between the base task and target task (Yosinki et al. 2014).

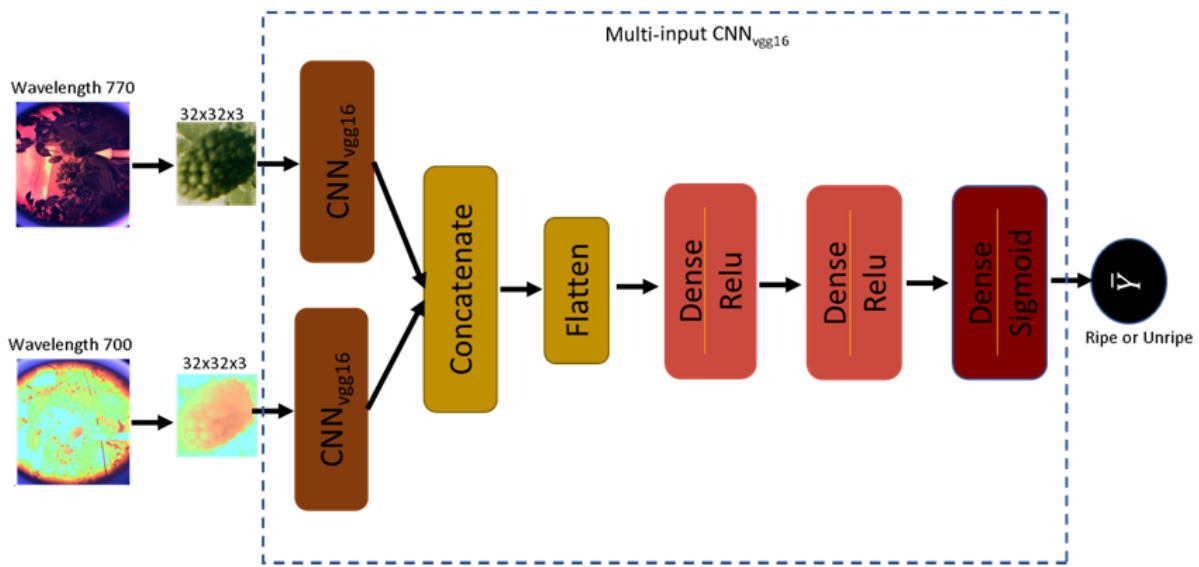

**Figure 3** The proposed multi-input $CNN_{vgg16}$ classifier for detecting ripeness in blackberry fruits. The input to the network is a pseudo-coloured 32x32x3 bispectral blackberry sample. This is best viewed in colour.

### 2.2.2 Ensemble Learning

With the best-performing hyperparameters, the multi-input $CNN_{vgg16}$ serves as the base model for ensemble learning and is henceforth named MCE. This paper creates homogeneous ensemble learners, that is, learners that share the same learning architecture as depicted in Figure 4. The concept of ensemble learning was first introduced in 1965 (Nilson, 1965) and has gained popularity in the classification task, see for instance (Liu et al., 22017; Shahhosseini et al., 2021).

Given blackberry dataset, $\vec{X}$, a bootstrap subset $\left(X^{*[1]}, X^{*[2]}, \cdots, X^{*[B]}\right)$ is created through bootstrap resampling (Efron, 1992). This method involves resampling (with replacement) from a sample dataset to create the bootstrap subsets. Since the base learners were not generated by cross-validation, the minority class of the blackberry dataset is oversampled using the process of random oversampling (Menardi and Torelli, 2014). Each bootstrap subset is trained using the base learner, multi-input $CNN_{vgg16}$, and is mathematically expressed as:

$$\hat{Y}_L^{*[i]} = \varphi\left(\sum_{j,k=1} w_{j,k}\left(x_{j,k}^{*[i]}\right) + b\right) \quad i = \{1,2,\cdots,B\}, j = \{1,2,\cdots,n\}, k = \{1,2,\cdots,n\} \tag{1}$$

where $x_{j,k}$ is the $j^{th}$ and $k^{th}$ vectorized bi-spectra samples from a bootstrap subset, $w_{j,k}$ represents the weights, $b$ is the bias and $\varphi$ is the activation function. Usually, the outcome, $\hat{Y}$, of the base learners, $L$, for $i^{th}$ bootstrap subset, should be combined into a single prediction result, hence it is important to consider the integration technique applied to the base learner's prediction.

Several base learners' integration techniques exist, including bagging, boosting, and stacking (Sewell, 2008). The bagging method assigns equal weight to all base learners' predictions, while the boosting method adjusts the weight of the training data of a prior base learner to improve on the prediction of a subsequent base learner. However, if a preceding base learner poorly performs, the performance of subsequent base learners cannot be guaranteed. Even though our multi-input $CNN_{vgg16}$ architecture has been optimized, making it appropriate for bootstrapping, the stacked generalization ensemble (SGE) framework, proposed by Wolpert in 1992 (Wolpert, 1992), is adopted. SGE combines the predictive accuracy of base learners to create a more powerful predictive ensemble learner. The SGE framework works by stacking the prediction results of base learners as feature vectors and uses a meta-learning algorithm to make a single prediction that is expected to outperform any of the base learner's predictions. The meta-learning algorithm used is the logistic regression algorithm, and the process is mathematically expressed as:

$$\hat{Y}_E = \frac{1}{1+e^{-(\beta_1 \hat{Y}_L^{*[i]} + \beta_0)}} \tag{2}$$

where $\hat{Y}^{*[i]}$ is a column feature vector with each column representing a base learner's, $L$ prediction, $\beta_1$ *is* the regression coefficient of the feature vectors estimated using a method generally known as maximum likelihood estimation, $\beta_0$ is the intercept of the feature vectors, and $\hat{Y}$ is the outcome of the ensemble learner, $E$.

## 3 Experimental Settings

### 3.1 Dataset

The blackberry fruit dataset is made up of blackberries that were assessed subjectively through fruit quality assessment (with and without an expert assessor – see below for more on this) for defining the ground truth labels of sample berries. The reason the expert assessors were involved at some point in the data collection was to be able to analyse the performance of machine sensory, MCE, in comparison to the human sensory. Therefore, the following question is posed: Can the proposed blackberry ripeness detection algorithm differentiate between a mature ripe and unripe berry with comparable performance to laboratory-level fruit quality assessment based on human sensory evaluation? If the proposed blackberry ripeness detection algorithm performs comparably in the field or outperforms laboratory-level fruit quality assessment, then blackberry growers will be able to overcome the need to transport blackberries to the laboratory and can minimize the number of mature unripe blackberries making their way to the market.

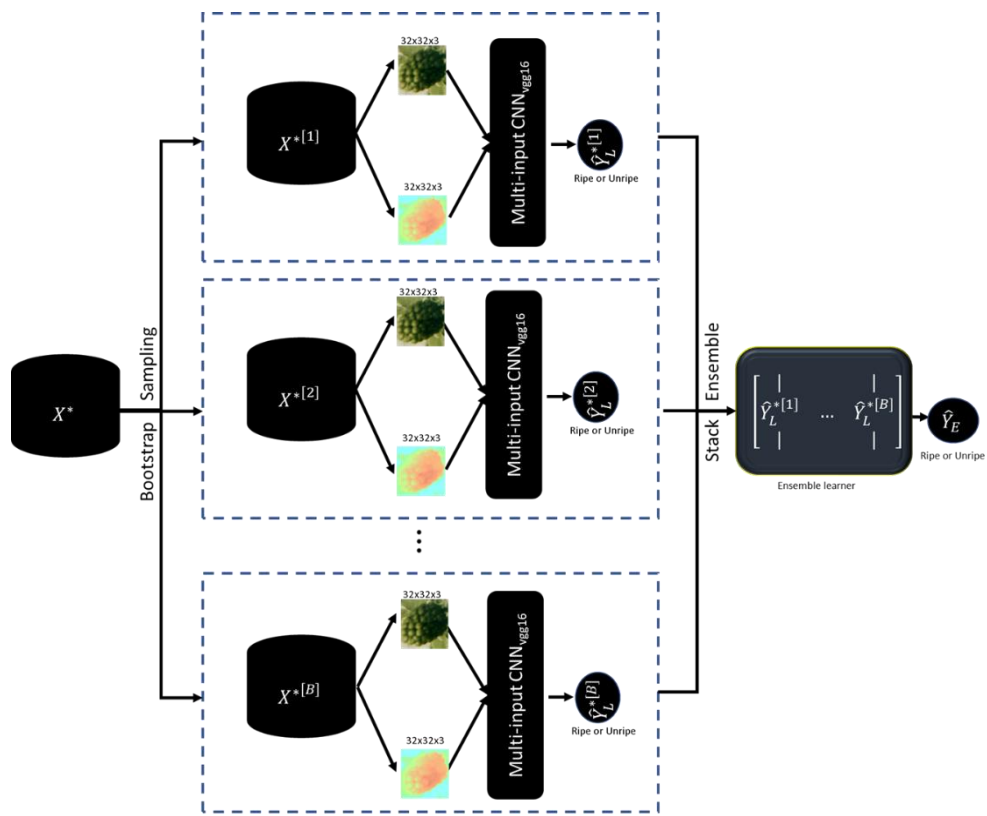

**Figure 4** Homogenous ensemble of learners with stacking architecture. Using the blackberry fruit dataset, a bootstrap subset is created through bootstrap re-sampling with the multi-input CNN$_{vgg16}$ as base learner. This is best viewed in colour.

In all, the blackberry fruit images in RGB and bispectral formats from Farm-A and Farm-B used for the experiments were labelled as, FA$_{RGB}$, FB$_{RGB}$, FA$_{bispec}$, and FB$_{bispec}$, with a total of 87, 97, 82, and 119 samples, respectively. Their combination is labelled as FA+B$_{RGB}$ and FA+B$_{bispec}$ with a total of 187 (148 ripe and 39 unripe) and 201 samples (164 ripe and 37 unripe), respectively, and split into 164 and 37 for the class of mature unripe berries. The bispectral and RGB sets were cropped to the blackberry of interest and resized. To improve the contrast of berries captured at a spectral wavelength of 770, the histogram equalization technique was applied. It should be noted that blackberry samples from Farm-A were assessed without fruit quality expert involvement, while blackberry samples from Farm-B were assessed by fruit quality assessment experts. The reason the total number of samples used for the experiment differs from the total number collected from both farms is due to the following:

i. A visible unripe appearance – images of a berry that show visible unripe attributes such as a presence of green or red colour were disregarded because such berries are not yet mature and therefore are outside the scope of the project.
ii. Motion blur – images were collected during the harvesting season and the interest was to capture berries with both the bispectral camera and the RGB camera working in the field in time before the pickers harvest them. As a result, some images were greatly affected by motion blur caused by camera motion. For this reason, these images were disregarded.
iii. No reference berry – the images where a pointer did not show a reference berry.

## 3.2 Classifier Evaluation and Optimization

The experimental scenarios for evaluating the proposed algorithm are comprehensively presented as follows: 1) classification experiments on FA$_{RGB}$, FB$_{RGB}$, FA$_{bispec}$, and FB$_{bispec}$, FA+B$_{RGB}$, and FA+B$_{bispec}$ using our multi-input CNN$_{vgg16}$ classifier without optimization, 2) Classification experiments on FA+B$_{bispec}$ using our multi-input CNN$_{vgg16}$ classifier with optimization, 3) Classification experiments on FA+B$_{bispec}$ using the optimized multi-input CNN$_{vgg16}$ with ensemble learning, and 4) proposed machine learning-based ripeness detector versus laboratory level fruit quality assessment using FB$_{bispec}$. For each scenario, the images were split into training, validation, and testing sets in the ratio of 60:20:20. The performance metrics used are weighted precision, weighted recall, weighted F1-score, and accuracy. Accuracy computes the ratio of correct prediction of mature ripe and unripe blackberries to the total number of mature blackberries. Weighted precision finds the ratio of mature ripe blackberries correctly predicted by computing the weighted percentage of true positives, (TP), for each class to the number of mature blackberries predicted, (TP), and false negative (FP), as mature ripe ones. Weighted recall finds the ratio of the number of mature ripe blackberries correctly predicted by computing the weighted percentage of true positives, (TP), for each class to the number of mature ripe blackberries comprising (TP) and false negatives (FN). The weighted F1-score is computed as the weighted average of weighted precision and weighted recall. These are mathematically expressed as follows.

$$Accuracy = \frac{TP+TN}{TP+FP+TN+FN} \quad (3)$$

$$Precision_w = \frac{TP_{weighted}}{TP+FP} \quad (4)$$

$$Recall_w = \frac{TP_{weighted}}{TP+FN} \quad (5)$$

$$F1-score_w = 2 * \left[\frac{(Recall_{weighted} * Precision_{weighted})}{Recall_{weighted} + Precision_{weighted}}\right] \quad (6)$$

The stratified random sampling method was employed to split $FA_{RGB}$, $FB_{RGB}$, $FA_{bispec}$, and $FB_{bispec}$, and $FA+B_{RGB}$, $FA+B_{bispec}$ into training, validation, and test sets. The reason the stratified random sampling cross-validation method was used was to address the imbalance problem between the mature ripe and unripe blackberry samples and to ensure that a balanced class ratio was maintained for each split. The optimization experiment employed the grid search algorithm for fine-tuning hyperparameters of the multi-input $CNN_{vgg16}$ classifier with the stratified $k$-fold cross-validation method. Given that the proposed classifier is of a multi-input structure, the grid search with a stratified $k$-fold cross-validation approach was implemented differently. For each hyperparameter, all other hyperparameters were unchanged, then a discrete grid with the hyperparameter values was set such that the stratified $k$-fold cross-validation method was implemented for each value of the grid. Using the stratified $k$-fold cross-validation method, the data was divided into $k$ subsets to ensure that the class ratio remained the same in each fold - a fold for testing and the remaining $k-1$ ($k$ =10) folds for training and validation. The performance of a hyperparameter was evaluated by averaging the results obtained in each $k-1$ experiment, then the process was repeated for a different value of the hyperparameter in the grid. In this paper, the fine-tuned hyperparameters are the number of neurons in the FC layer, optimization functions, batch size, and epoch. The descriptions and value settings of the hyperparameters as used in the experimental scenario (1) for the combined dataset are presented in Table 2, while the hyperparameters for the experiment scenario (2) are provided alongside their results in Table 4.

**Table 1** Evaluation of different CNN models for blackberry ripeness detection.

| Model | Depth | Image Size | FC Layer | Precision | Recall | F1-Score | Test Accuracy |
|---|---|---|---|---|---|---|---|
| DenseNet | 242 | 32x32x3 | $N^{512,256}$ | 0.872 | 0.735 | 0.772 | 0.878 |
| InceptionV3 | 189 | 75x75x3 | $N^{256,64}$ | 0.808 | 0.829 | 0.797 | 0.829 |
| Xception | 81 | 71x71x3 | $N^{512,256}$ | 0.777 | 0.683 | 0.712 | 0.683 |
| MobileNet | 55 | 75x75x3 | $N^{512,256}$ | 0.873 | 0.878 | 0.875 | 0.878 |
| **Vgg16** | **16** | **32x32x3** | $\mathbf{N^{1024,1024}}$ | **0.898** | **0.902** | **0.897** | **0.902** |

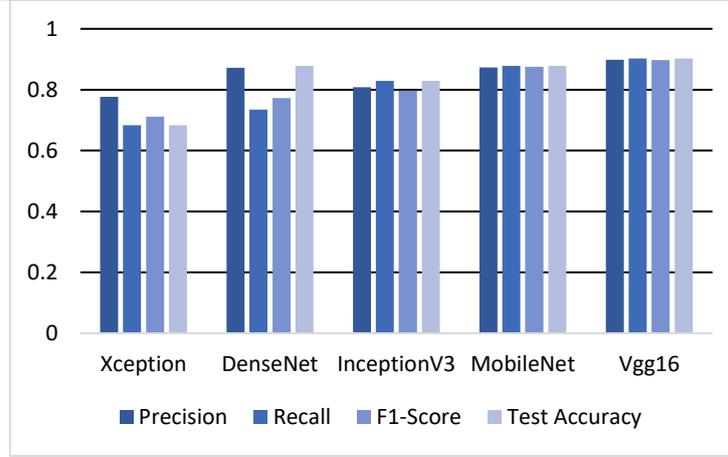

**Figure 5** Comparison of pre-trained CNN models performance on blackberry ripeness detection. The VGG16 emerged the best.

**Table 2** The multi-input $CNN_{vgg16}$ classifier hyperparameters

| Hyperparameter | Description | Initial Values |
|---|---|---|
| FC Layer | The fully connected (FC) layer defines the number of neurons connected to each input from the last convolutional layer of the pre-trained VGG16 network for the weights of the target task to be learned. | $N^{1024,1024}$ |
| Optimization Function | An algorithm that minimizes the loss function of the network during training. | Adam |
| Batch Size | The number of samples utilized in one iteration. | 10 |
| Epoch | Defines the number of passes made through the entire training dataset during training. | 20 |

## 4. Experimental Results and Discussion

Experiments were carried out in the order of the experimental scenarios and therefore results, and discussions, are reported accordingly.

### 4.1 Multi-input $CNN_{vgg16}$ classifier without optimization

The results of applying the multi-input $CNN_{vgg16}$ classifier on the $FA_{RGB}$, $FB_{RGB}$, $FA_{bispec}$, $FB_{bispec}$, $FA+B_{RGB}$, and $FA+B_{bispec}$, respectively, are presented in Table 3 and Figure 6. It can be observed in Table 3 and Figure 6 that the multi-input $CNN_{vgg16}$ classifier which takes as input the bispectral images of a blackberry fruit acquired at two spectral wavelengths, 700 and 770 nm, can detect the ripeness of mature blackberry fruits than the RGB images of both Farm-A and Farm-B. The average level performance of the RGB-based model for both individual and combined Farm-A and Farm-B datasets is not surprising because mature ripe and unripe blackberry fruits are not visually discernible from each other.

The multi-input $CNN_{vgg16}$ model can be observed in Table 3 to overfit to $FA_{RGB}$ and $FB_{RGB}$ training sets, but it appeared to be addressed with more examples of ripe and unripe class when the combined $FA+B_{RGB}$ data set was used. However, accuracy remained a little above average and skewed toward the ripe class. On the other hand, a stable performance was achieved with $FA_{bispec}$ and $FB_{bispec}$ data sets. The 6.9% increase in accuracy achieved with the $FB_{bispec}$ set might be because the samples from Farm-B were subjectively assessed by expert fruit assessors and therefore were more correctly labelled.

**Table 3** The ripeness detection performance of the multi-input $CNN_{vgg16}$ classifier on different farm datasets.

| Data Set | Precision | Recall | F1-Score | Train Accuracy | Test Accuracy |
|---|---|---|---|---|---|
| $FA_{rgb}$ | 0.556 | 0.556 | 0.556 | 0.783 | 0.556 |
| $FB_{rgb}$ | 0.700 | 0.700 | 0.700 | 0.974 | 0.699 |
| $FA_{bispec}$ | 0.882 | 0.824 | 0.827 | 0.985 | 0.824 |
| $FB_{bispec}$ | 0.889 | 0.893 | 0.889 | 0.938 | 0.893 |
| $FA+B_{rgb}$ | 0.623 | 0.789 | 0.696 | 0.792 | 0.789 |
| $FA+B_{bispec}$ | **0.898** | **0.902** | **0.897** | **0.963** | **0.902** |

With the bispectral combined dataset, $FA+B_{bispec}$, the multi-input $CNN_{vgg16}$ achieved 90.2% accuracy. However, precision and recall present interesting results – 90.2% and 89.8% accuracies, respectively. Given that the mature unripe berries constitute about 20% of the set and the remaining 80% of the mature ripe berries, both metrics are useful for evaluating the model's performance. The precision gives more insight into the performance of the model on the unripe class, while the recall on the ripe class. A 90.2% recall shows the model correctly classified the ripe class, and a precision value of 89% shows the model performed equally well on the unripe class. It should be noted that the true class was assigned to the unripe blackberry fruits and the ripe blackberry fruits, the negative class. This ordering was automatically defined by the order of the sample folders, but does not in any way affect the result of the model. Overall, it can be concluded that the spectral filters of wavelengths, 700 nm, and 770 nm, were able to uncover the hidden features of mature berries that can be used to discriminate between a ripe and unripe blackberry fruit.

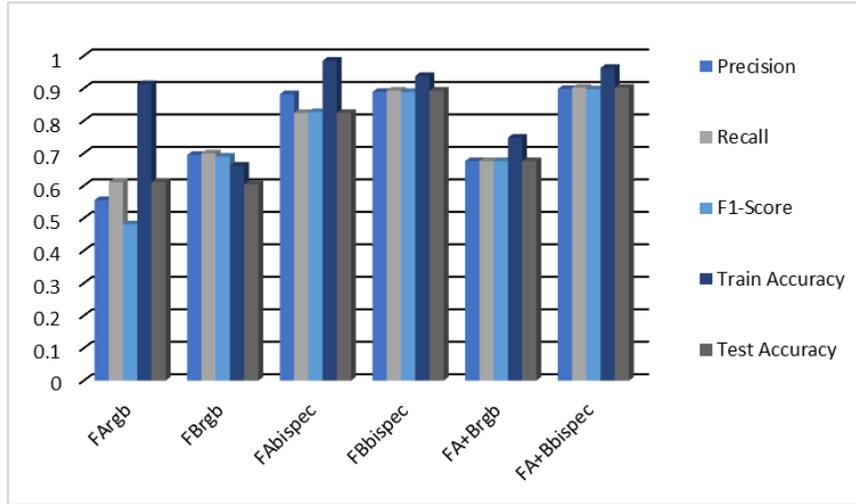

**Figure 6** The Multi-input $CNN_{vgg16}$ performance on different farms, Farm-A and Farm-B, datasets.

## 4.2 Multi-input $CNN_{vgg16}$ classifier with optimization

The multi-input $CNN_{vgg16}$ hyperparameters were optimized through grid search and stratified k-fold cross-validation. The hyperparameter, its corresponding values, and performance across different metrics are presented in Table 4 and discussed in subsequent paragraphs. It should be noted that the best values of each hyperparameter based on the performance metrics, precision, recall, and F1-score, become the input to the multi-input $CNN_{vgg16}$ for computing the next hyperparameter.

To determine the best hyperparameters for the FC layers of the multi-input $CNN_{vgg16}$ classifier, the experiments included fine-tuning the layers by 1) to have the same number of neurons, and 2) to vary the first FC layer while the second layer is kept constant. It was interesting to observe from the experiments that expanding the feature space at the FC layer to twice its input size increased the proposed model's discriminative capability. The best results were recorded across the performance metrics, as highlighted in Table 4, of train accuracy, test accuracy, precision, recall, and F1-score with the design. The obvious logical explanation is that it helps the input to the FC layer to be learned by more weights, which increases the significance of relevant features. However, when the first FC layer equalled the size of the input to the layer, and the second FC layer was reduced by half its size, higher results were observed across all the performance metrics. By comparison, their precision and test accuracies are the twice-in-size FC layer achieved 92.1% and 91.5%, respectively whereas, the varying-size FC layer achieved 92.3% and 91.7%. These are a difference of 0.2% which is a minimal difference, but it means that the varying-size FC layer will feed into the multi-input $CNN_{vgg16}$ architecture for finding the next best hyperparameter.

The next is to determine the best optimization function for the multi-input $CNN_{vgg16}$. The optimization functions compared are Adam optimizer, stochastic gradient descent, and adaptive gradient optimization functions with their hyperparameters' setup as follows: learning rate of 0.01, 0.01, 0.001, and 0.001 for SGD (Sutskever et al., 2013), Adam (Kingma and Ba, 2014), and AdaGrad (Duchi et al. 2011), respectively. Adam being a second-order optimizer uses two exponential decay rates set at 0.9 and 0.999 for the first and second moments, respectively. The Adam optimizer performed better in comparison to the SGD and AdaGrad optimizers, with a precision rate of 91.50%. The AdaGrad had the least performance, which was expected because it is particularly suited to sparse data.

Using the Adam optimizer, the values of 16 and 40 for batch size and epoch, respectively, were determined to be the values to best update the weights of the network towards a global minimum given that their precision and test accuracies of 91.80% and 95.70%, and 91.70% and 91.90%, respectively, are the highest of the scores.

### 4.3 Ensemble learning with optimized multi-input $CNN_{vgg16}$

The MCE model performance can be observed in Figure 7, Figure 8 and Table 5. The MCE model achieved a 95.10% accuracy on the test set, as can be observed in Table 5, and it is a 2.8% increase in performance to the optimized multi-input $CNN_{vgg16}$ model. Using area under the curve (AUC) and precision-recall curve as shown in Figure 7, the performance of MCE under class imbalance further confirms that the stratified random sampling cross-validation used for building the base model addressed the class imbalance. Experiment is extended to investigate the impact of unforeseen on-farm environmental factors on MCE model. This is motivated by 1) the varying view angles of the blackberry fruit during harvesting, 2) change in fruit picker viewpoint relative to the fruit position, and 3) changes in environment lighting due to weather conditions (cloudy, sunny, or rainy) or time of day. These factors were modelled using augmentation techniques such as rotation, zoom, and brightness and are significant towards validating the model's capability to withstand unforeseen environmental factors. With a 10 degrees rotation, 0.2 to 1.0 zoom values, and 0.2 to 0.1 brightness values, MCE achieved a performance accuracy of 90.2% as can be observed in Figure 8. Even though the complexity of the environmental factors decreased the accuracy of the model, it is still above 90%.

Further evaluation includes visualising MCE performance using the confusion matrix. Likewise, the model's performance in the presence of environmental factors, as investigated using the augmentation techniques, is visualised using the confusion matrix. The MCE without augmentation is presented in Figure 9 (a) and with augmentation in Figure 9 (b-c). The confusion matrix reveals the impact the environmental factors have on the proposed model's performance.

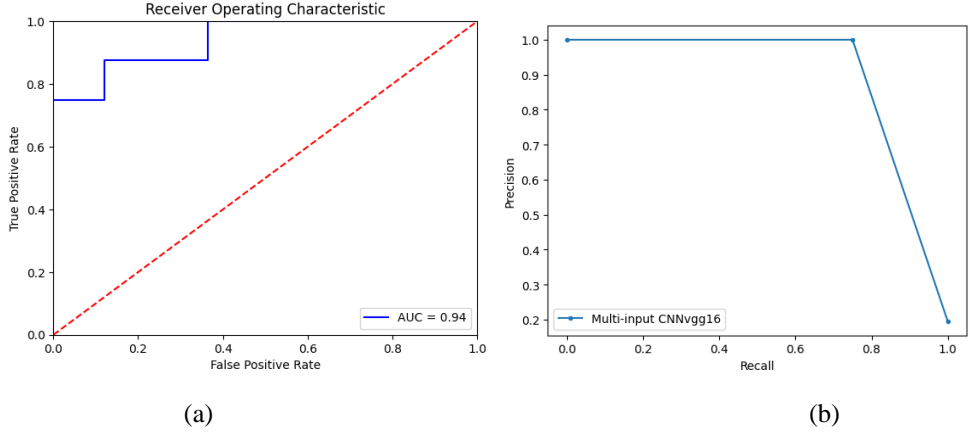

| (a) | (b) |

Figure 7 Evaluation of MCE performance under class imbalance using AUC (a) and precision-recall curve (b).

Table 4 The Multi-input CNNvgg16 hyperparameter optimization evaluation. The best hyperparameter feeds into the next experiment.

| Hyperparameter | Value | Precision | Recall | F1-Score | Train Accuracy | Test Accuracy |
|---|---|---|---|---|---|---|
| FC Layer | $N^{256,256}$ | 0.910±0.048 | 0.905±0.051 | 0.893±0.061 | 0.959±0.039 | 0.905±0.048 |
| | $N^{512,512}$ | 0.913±0.044 | 0.900±0.056 | 0.888±0.064 | 0.961±0.032 | 0.900±0.053 |
| | $N^{1024,1024}$ | **0.921±0.041** | **0.917±0.038** | **0.911±0.043** | **0.968±0.029** | **0.915±0.049** |
| | $N^{2048,2048}$ | 0.920±0.035 | 0.909±0.042 | 0.899±0.056 | 0.964±0.024 | 0.895±0.048 |
| | $N^{256,256}$ | 0.883±0.099 | 0.898±0.060 | 0.882±0.079 | 0.961±0.048 | 0.898±0.058 |
| | $N^{512,256}$ | **0.923±0.048** | **0.922±0.043** | **0.914±0.048** | **0.973±0.017** | **0.917±0.037** |
| | $N^{1024,256}$ | 0.887±0.096 | 0.900±0.052 | 0.887±0.073 | 0.963±0.049 | 0.888±0.043 |
| | $N^{2048,256}$ | 0.876±0.087 | 0.887±0.045 | 0.873±0.064 | 0.961±0.048 | 0.900±0.039 |
| Optimization Function | Adam | **0.915±0.056** | **0.913±0.054** | **0.910±0.034** | **0.970±0.029** | **0.915±0.060** |
| | SGD | 0.872±0.084 | 0.876±0.044 | 0.845±0.069 | 0.927±0.043 | 0.876±0.041 |
| | AdaGrad | 0.799±0.105 | 0.826±0.017 | 0.764±0.037 | 0.861±0.022 | 0.827±0.017 |
| Batch Size | 4 | 0.909±0.049 | 0.907±0.044 | 0.899±0.049 | 0.972±0.018 | 0.907±0.042 |
| | 6 | 0.903±0.049 | 0.902±0.047 | 0.893±0.056 | 0.963±0.033 | 0.902±0.045 |
| | 8 | 0.910±0.052 | 0.907±0.049 | 0.897±0.058 | 0.966±0.038 | 0.807±0.047 |
| | 10 | 0.902±0.058 | 0.897±0.057 | 0.890±0.067 | 0.966±0.028 | 0.897±0.053 |
| | 12 | 0.909±0.054 | 0.905±0.051 | 0.893±0.059 | 0.963±0.031 | 0.910±0.048 |
| | 14 | 0.903±0.054 | 0.900±0.053 | 0.888±0.063 | 0.965±0.032 | 0.900±0.051 |
| | 16 | **0.918±0.056** | **0.917±0.054** | **0.911±0.057** | **0.975±0.018** | **0.917±0.051** |
| Epoch | 20 | 0.909±0.053 | 0.907±0.047 | 0.903±0.050 | 0.971±0.032 | 0.902±0.048 |
| | 40 | **0.957±0.026** | **0.954±0.029** | **0.949±0.036** | **0.980±0.016** | **0.919±0.046** |
| | 60 | 0.898±0.039 | 0.893±0.035 | 0.879±0.043 | 0.961±0.029 | 0.893±0.033 |
| | 80 | 0.955±0.032 | 0.951±0.035 | 0.945±0.042 | 0.948±0.058 | 0.885±0.053 |

|  | 100 | 0.923±0.054 | 0.919±0.048 | 0.914±0.078 | 0.964±0.028 | 0.907±0.032 |

**Table 5** Experimental results of the multi-input CNN$_{vgg16}$ ensemble learner with on-farm environmental factors

| State | Precision | Recall | F1-Score | Test Accuracy |
|---|---|---|---|---|
| No augmentation | 0.954 | 0.951 | 0.948 | 0.951 |
| Rotation+Zoom | 0.933 | 0.927 | 0.919 | 0.927 |
| Rotation+Zoom+Brightness | 0.913 | 0.902 | 0.889 | 0.902 |

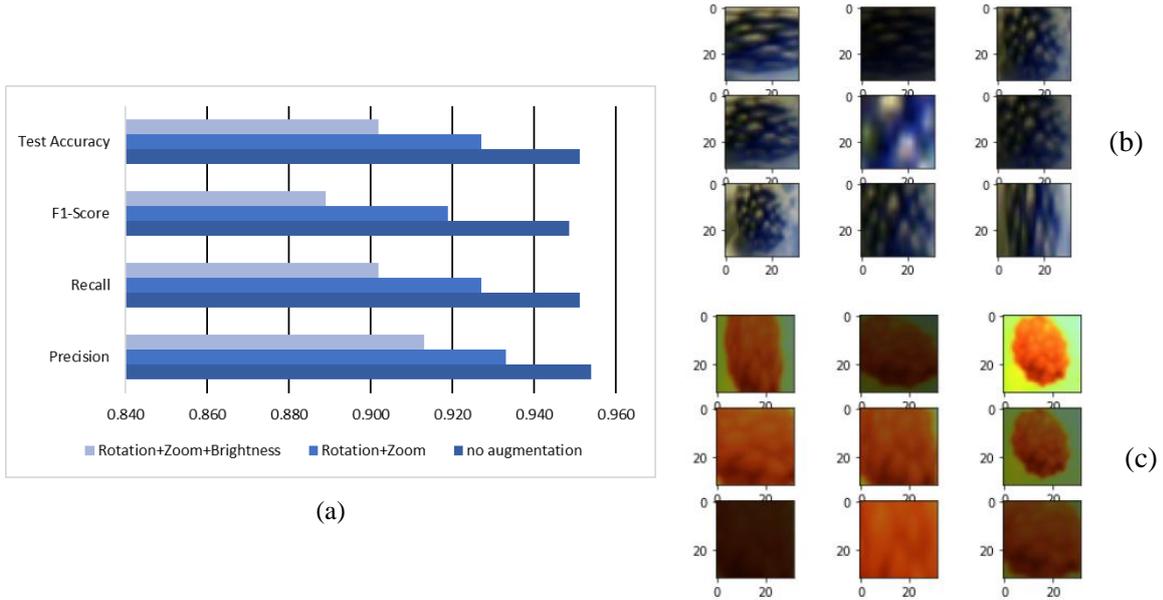

**Figure 8** The MCE performance without and with augmentation. The rotation, zoom and brightness transformations test model capability to withstand unforeseen challenges.

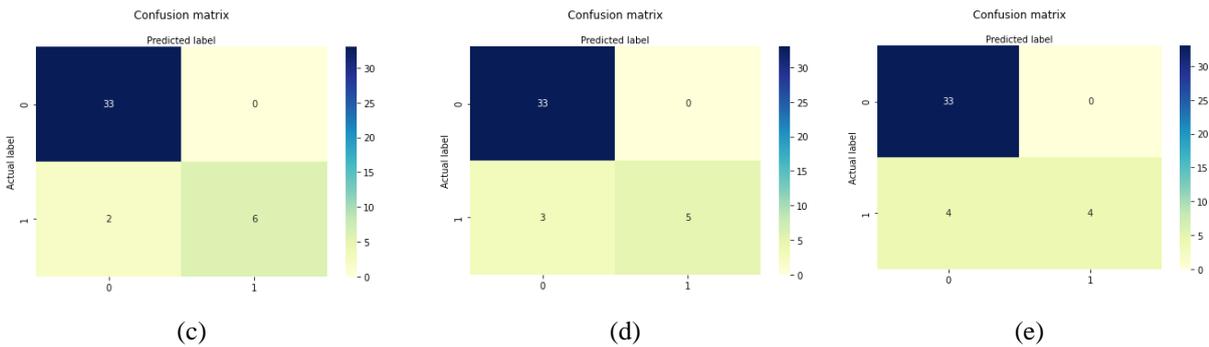

**Figure 9** The confusion matrix of MCE using Farm A+B$_{bispec}$ dataset. (a) no augmentation (b) with rotation and zoom augmentation, and (c) with rotation, zoom, and brightness augmentation.

## 4.4 Human ripeness detection vs machine ripeness detection

Humans have evolved very sensitive and specialized sensory organs, which give them the ability to test fruit quality through a process called sensory evaluation. This innate ability can be further trained, and detailed sensory evaluation can be carried out by expert panels who are adept at using sensory attributes such as odour, flavour, taste, and texture/appearance to describe and quantify fruit quality attributes. Typically, most blackberry fruit growers would subject some of the harvested berries in any given pick throughout the season to some degree of organoleptic evaluation to ensure that the berries meet consumer requirements. Additionally, Brix is usually measured internally and by or at the request of retailers. In this paper, expert sensory evaluations were carried out for berries sampled from Farm-B. The evaluation involved a trained evaluator who measured berry fruit quality using sensory attributes and a scoring system as follows: shininess (1=dull, 5=shiny), uniformity of colour (1=uneven, 5 =even), firmness (1=soft, 5=firm), skin strength (berry skin broken after manipulation =Y/N), flavor (1=aromatic, 5=poor), sweetness (1=sweet, 5=sour), texture (1=melting, 5=crunchy), and human colour ripeness (0=green, 1= red, 2= black but underripe, 3= fully ripe, 4=overripe). Based on these factors, the human ripeness decisions are recorded as *human ripeness* in Table 6.

The machine does not have all the sensory organs as the human, but it tries to compensate for it by maximizing what resembles the visual senses using MCE. This is to enable it to learn characteristics of the blackberry fruit such as points, lines, edges, textures, shapes, and a combination of other factors that correlates to ripeness. To be able to compare machine sensory to the human, the confidence score obtained through evaluating the MCE was used because it measures how confident the model is in its prediction of a given class. There are two classes, ripe and unripe, which are represented as 0 and 1. Each blackberry fruit was labelled under *target* according to *human ripeness,* as given in Table 6.

To qualitatively analyse the MCE performance, the Pearson correlation matrix is observed for MCE confidence against the human sensory variables. As can be observed in Figure 10, it is evident that the machine sensory is highly and positively correlated to the human sensory evaluation over blackberry fruit texture. A detailed analysis, using Table 6, shows that the texture scores of berries vary with ripeness levels. For example, the machine reached a 99% confidence score for an unripe berry with a texture score of 5 and a ripeness level of 1.5. Additionally, the human classified one of two unripe berries with texture scores of 4, to be ripe and the other unripe. However, the machine considered both berries to be of different unripe degrees with confidence scores of 78.48%, and 96.85%. Machines are designed to complement human efforts in automation and outperform humans in a task requiring efficiency. The MCE model demonstrated this capability in two ways. Berries *3D9262* and *976122* were labelled unripe based on their human ripeness scores, however, the machine considered them to be near-ripe by confidence scores of 2.49% and 1.09%, respectively. Another scenario where the human scored a berry, *7DE558*, as ripe, the machine classified it to be unripe with a 96.85% confidence score.

Table 6 Comparison of human sensory and machine sensory for blackberry fruit ripeness detection.

| Berry ID | Mass (g) | Shininess | Colour Uniformity | Firmness | Skin strength | Flavor | Sweetness | Texture | Human Ripeness | Target | Machine Ripeness | Confidence (%) |
|---|---|---|---|---|---|---|---|---|---|---|---|---|
| 3D9262 | 8.6849 | 5 | 4 | 4 | N | 2 | 3 | 3 | 2.5 | 1 | 0 | 2.49 |
| 976122 | 6.4945 | 4 | 5 | 4 | N | 4 | 4 | 3 | 2.5 | 1 | 0 | 1.09 |
| F3B65F | 7.9683 | 3 | 4 | 4 | N | 5 | 5 | 4 | 2 | 1 | 1 | 78.48 |
| 061AE0 | 4.9857 | 5 | 4 | 5 | N | - | - | - | 2 | 1 | 1 | 95.72 |
| 7DE558 | 6.6189 | 4 | 5 | 4 | N | 4 | 4 | 4 | 3 | 0 | 1 | 96.85 |
| A2761F | 7.0942 | 4 | 5 | 4 | Y | 3 | 3 | 3 | 3 | 0 | 0 | 0.19 |
| 8B49E5 | 8.1658 | 2 | 4 | 2 | N | 2 | 3 | 2 | 3 | 0 | 0 | 0.01 |
| 165B41 | 6.3459 | 3 | 1 | 4 | N | 5 | 5 | 5 | 1.5 | 1 | 1 | 99.99 |
| FF9EA0 | 5.9842 | 4 | 1 | 4 | N | 5 | 5 | 5 | 1.5 | 1 | 1 | 99.86 |
| E206D2 | 11.2435 | 4 | 5 | 3 | N | 2 | 2 | 2 | 3.5 | 0 | 0 | 0.08 |

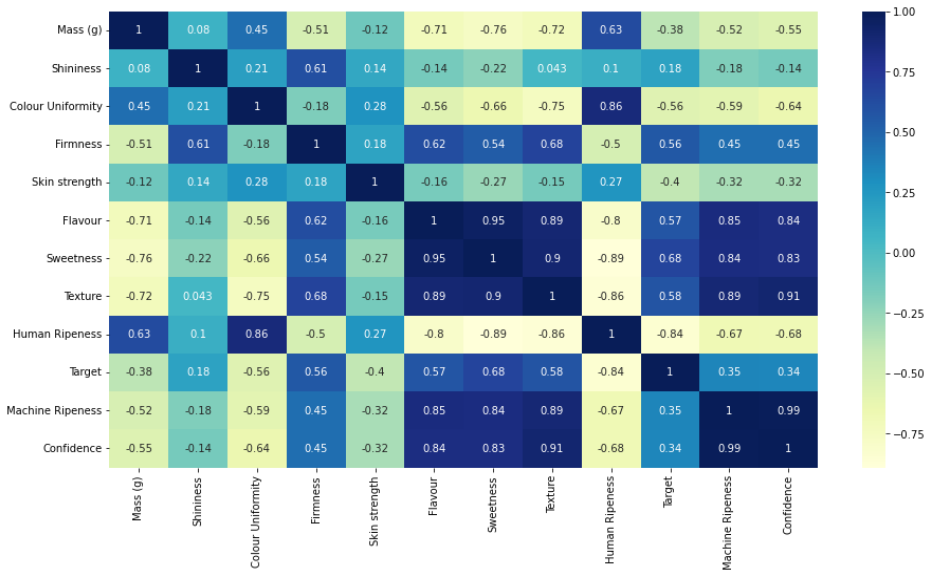

**Figure 10** Qualitative analysis of the confidence of the machine sensory and human sensory using Pearson correlation matrix.

## 5. Conclusion

The traits of ripeness of mature blackberry (Rubus fruticosus) fruits are very subtle and renders the task of visual assessment of mature blackberry difficult for growers and pickers. Therefore, this paper proposed a novel multi-input CNN designed from a VGG16 model pre-trained on the ImageNet dataset, and the fully connected layers were optimized for learning and served as the base model for ensemble learning. The ensemble learning comprised homogeneous ensemble learners and their predictions were ensembled using the SGE framework for meta-learning in order to classify a blackberry fruit image as ripe or unripe. The input to the classifier was the blackberry fruits acquired from a stereo sensor with Vis-NIR spectral filters at wavelengths of 700 nm and 770 nm. This is the first research to investigate blackberry fruit ripeness, and the novel multi-input CNN ensemble classifier proposed in this paper extends the possibilities of assessing fruit ripeness using hyperspectral images instead of their indices. Through experiments, the proposed model achieved 95.1% accuracy on unseen sets and with increased environmental challenges achieved 90.2% accuracy. These are an 11.3% and 16.2% increase in accuracy compared to the RGB-based model. Further experiments compared machine ripeness sensing capability with the human sensory ability, and it was interesting to observe that the machine sensory was positively correlated to the human sensory evaluation of ripeness over blackberry fruit. It was also efficient in overcoming the fuzzy limitations of the human sensory. However, since the hyperspectral imagery in this paper was laboratory-driven, the variability in spectral reflectance due to environmental factors was not considered. Meanwhile, it will be worth investigating the interaction of the environmental factors on reflectance spectra in on-farm settings. Further consideration for future work includes enhancing the stereo camera lens based on the outcome of the on-farm hyperspectral imaging and variability impact analysis. This will likely increase the possibilities of generating numerous samples of better-quality images in a more efficient way, thereby increases the robustness of the proposed model to on-farm spectral variability. There is also room to expand the samples to include the overripe cases which extends the work beyond a binary classification task. With an end goal to develop an ergonomic non-destructive tool for on-farm fruit ripeness assessment, the study will investigate the applicability of the proposed system in a robot picker or augmented reality headset for a human picker.

**Acknowledgement**

Special thanks to Windmill Hill Fruits Ltd in Herefordshire, UK, and Clockhouse Farm in Kent, UK., for granting access to their farms and permitting the sampling of their blackberry fruits.

**Funding**

This work was supported by Innovate UK (50877). The funding sources had no role in the study design, execution, analysis and interpretation of the data, or writing of this manuscript.

**Conflicts of interest**

The authors declare that there is no conflict of interest.

**Authors Contributions**

Conceptualization, funding acquisition, and project supervision, B.T. and M.S.; grower challenges, farm coordination, and harvest seasons, H.D.; literature survey and project scope, M.S.; Literature review, methodology, coding, validation, testing, and manuscript – original draft, C.O.; hyperspectral image wavelength selection, B.L.; stereo camera design and data collection, B.T. and B.W.; hyperspectral imaging fruit ripeness laboratory assessment, C.W. and F.F.F; All authors reviewed, edited and approved the final manuscript.